\documentclass[journal]{IEEEtran}

\usepackage{amsmath,amssymb}

\usepackage{graphicx}
\usepackage{subcaption}
\usepackage[font=footnotesize,labelfont=bf]{caption}
\usepackage{array}
\usepackage{booktabs}
\usepackage{tabularx}
\usepackage{multirow}
\usepackage{adjustbox}
\usepackage{hyperref}

\usepackage{tikz}
\usetikzlibrary{arrows.meta,positioning,fit,backgrounds,calc}
\usepackage{xcolor}
\usepackage{graphicx}
\usepackage{newtxtext,newtxmath} 

\usepackage{cite}
\usepackage{url}

\usepackage{xcolor}
\usepackage{tikz}
\usetikzlibrary{arrows.meta,positioning}

\usepackage{algorithm}
\usepackage{algorithmic}



\begin{document}

\title{JEPA for AI-Native 6G:\\Predictive Representations and Open Challenges}

\author{Sheikh~Salman~Hassan,~\IEEEmembership{Member,~IEEE},
        Irshad~A.~Meer,~\IEEEmembership{Member,~IEEE},
        Almoatssimbillah~Saifaldawla,~\IEEEmembership{Member,~IEEE},
        Yan~Kyaw~Tun,~\IEEEmembership{Senior Member,~IEEE},
        Mustafa~Ozger,~\IEEEmembership{Member,~IEEE}{,}
        Madyan~Alsenwi,~\IEEEmembership{Member,~IEEE}{,}
        Nguyen~Van~Huynh,~\IEEEmembership{Member,~IEEE},
        Woong-Hee~Lee,
        Cedomir~Stefanovic{,}~\IEEEmembership{Senior Member,~IEEE},
        Mathini~Sellathurai,~\IEEEmembership{Fellow,~IEEE},\\
        Henk~Wymeersch,~\IEEEmembership{Fellow,~IEEE},
    and~Tharmalingam~Ratnarajah,~\IEEEmembership{Senior~Member,~IEEE}%
\thanks{%
\IEEEcompsocthanksitem S.~S.~Hassan is with IDCoM, University of Edinburgh,
Edinburgh EH9 3BF, UK. E-mail: shassan@ed.ac.uk.
\IEEEcompsocthanksitem I.~A.~Meer is with KTH Royal Institute of
Technology, SE-100 44 Stockholm, Sweden. E-mail: iameer@kth.se.
\IEEEcompsocthanksitem A.~Saifaldawla and M.~Alsenwi are with the
SnT, University of Luxembourg, L-1855 Luxembourg.
E-mails: moatssim.saifaldawla@uni.lu, madyan.alsenwi@ieee.org.
\IEEEcompsocthanksitem Y.~K.~Tun, M.~Ozger, and C. Stefanovic are with the Department of Electronic Systems, Aalborg University, 9220 Aalborg, Denmark. E-mails: ykt@es.aau.dk, mozger@es.aau.dk, cs@es.aau.dk.
\IEEEcompsocthanksitem N.~V.~Huynh is with the School of Computer Science and Informatics, University of Liverpool, Liverpool L69 3DR, UK. E-mail: huynh.nguyen@liverpool.ac.uk.
\IEEEcompsocthanksitem W.-H.~Lee is with the Division of Electronics and Electrical Engineering, Dongguk University, Seoul 04620, Republic of Korea. Email: woongheelee@dongguk.edu.
\IEEEcompsocthanksitem M.~Sellathurai is with the School of Engineering and Physical Sciences, Heriot-Watt University, Edinburgh EH14 4AS, UK. E-mail: M.Sellathurai@hw.ac.uk.
\IEEEcompsocthanksitem H. Wymeersch is with the Department of Electrical Engineering, Chalmers University of Technology, 41296 Göteborg, Sweden. E-mail: henkw@chalmers.se.
\IEEEcompsocthanksitem T.~Ratnarajah is with the Department of Electrical
and Computer Engineering, San Diego State University,
San Diego, CA 92182, USA. E-mail: t.ratnarajah@ieee.org.%
}
}

\maketitle

\begin{abstract}
Sixth-generation (6G) networks are moving toward AI-native operation, where learning modules are embedded across the radio access network (RAN), edge, and core. This transition requires learning from limited labels, heterogeneous wireless and network data, partial observations, non-stationary propagation, and latency-constrained control loops. Joint-embedding predictive architecture (JEPA) is a promising self-supervised paradigm for this setting because it predicts missing or future representations in latent space instead of reconstructing raw measurements or using contrastive negative samples. This article presents a wireless-oriented tutorial on JEPA for 6G intelligence. We define the JEPA training mechanism, describe how CSI, beam measurements, KPIs, topology graphs, and sensing observations can be tokenized and masked, and position the learned encoder as a predictive representation layer for RAN, O-RAN, edge, and core functions, with task-specific heads or controllers producing final decisions. Then we present an illustrative, beam-management case study suggesting that a wireless-aware target, specifically an auxiliary future beam-energy target during self-supervised pretraining, can improve label efficiency and robustness across shifted deployment conditions relative to a supervised source domain. Finally, we outline open challenges in multi-timescale prediction, action-conditioned modeling, distributed training, trustworthiness, efficient deployment, benchmarking, and standardization.
\end{abstract}


\section{Introduction \& Background}

6G networks are expected to move beyond improvements in throughput, latency, connectivity, and reliability toward artificial intelligence (AI)-native operation, where learning modules are embedded across the radio access network (RAN), edge, and core \cite{8808168}. In current 5G systems, AI is mainly used as an optimization or analytics layer for traffic prediction, resource allocation, anomaly detection, and network management. For example, the network data analytics function provides analytics to other network functions upon request \cite{10578162}. In parallel, 3GPP has started studying AI closer to the radio interface, where TR~38.843 investigates AI for the 5G New Radio (NR) air interface, with initial focus on channel state information (CSI) feedback, beam management, and positioning \cite{3gpp_tr38843}. These efforts show that AI is moving from high-level analytics toward radio-interface functions in 5G-Advanced \cite{majumdar2025aintdai}.

The 6G vision extends through integrated sensing and communications (ISAC), non-terrestrial networks, ultra-dense edge intelligence, semantic communication, digital twins, and mission-critical control \cite{8869705}. These services require network intelligence to process heterogeneous observations distributed across users, base stations, edge servers, and core functions. These observations differ in format, sampling rate, reliability, and meaning, and may arrive asynchronously or be missing in some network segments. This creates a multimodal learning problem involving wireless and network modalities such as CSI, I/Q samples, beam measurements, RAN key performance indicators (KPIs), topology information, mobility traces, traffic telemetry, and sensing observations. The challenge is further amplified by terrestrial links with non-terrestrial networking dynamics, near-real-time (RT) O-RAN latency constraints, and multi-vendor model lifecycle requirements.

Supervised learning has shown strong performance in wireless tasks such as modulation recognition, channel estimation, beam prediction, traffic classification, and interference detection \cite{9058656}. However, supervised models require labeled data, which is sensitive to domain shifts across sites, frequency bands, antenna configurations, hardware platforms, mobility patterns, and interference conditions. These limitations are critical for 6G because labels are expensive, network conditions change continuously, and the relevant state is only partially observed by any single entity. Therefore, maintaining a separate supervised model for every task, deployment, and operating condition is difficult in an AI-native architecture.

Self-supervised learning (SSL) can exploit unlabeled operational data generated by the network \cite{10930396}. Existing wireless SSL methods are often based on reconstruction or contrastive learning. Reconstruction-based methods, such as masked autoencoders (MAEs), can be effective when the target is carefully designed, but may spend capacity on raw measurement details, hardware artifacts, or noise that are not directly useful for control \cite{guler2025multitask, guler2026pilotwimae}. Contrastive methods can learn robust representations, but depend on task- and channel-specific augmentations and negative-pair construction \cite{11071965}. Thus, reconstruction and contrastive SSL motivate a complementary paradigm focused on predictive latent network-state modeling.

Joint-embedding predictive architecture (JEPA) is a paradigm in which, rather than reconstructing raw signals or contrasting augmented samples, the model predicts latent representations of missing or future observations from the available context \cite{Assran_2023_CVPR}. This principle aligns with AI-native 6G, where the goal is often to infer hidden radio environment and network states useful for future decisions rather than to classify an isolated snapshot. A JEPA encoder can summarize how beam quality, link reliability, traffic demand, or anomalous behavior may evolve. In this way, JEPA can complement 3GPP-aligned AI use cases such as CSI feedback, beam management, and positioning, while also extending to ISAC, O-RAN automation, and non-terrestrial networks.

JEPA can serve as a predictive representation layer for AI-native 6G networks, but its effectiveness depends on wireless-aware design choices. These include encoder architecture, context and target views, masking patterns, prediction horizons, and downstream task heads. Therefore, adapting JEPA to 6G requires aligning latent prediction targets with wireless control objectives. This article provides a tutorial and architectural perspective on JEPA for 6G wireless and network data. We translate JEPA into a wireless workflow, including how heterogeneous observations are tokenized, how context and target views are constructed, what is trained during SSL pretraining, what is reused after pretraining, and how downstream heads or controllers consume the learned representation. Since JEPA produces predictive embeddings, final decisions such as beam choices, modulation-and-coding-scheme (MCS) selections, handover actions, slice policies, or anomaly alarms are produced by task-specific modules built on top of the encoder. The main contributions are summarized as follows:
\begin{itemize}
    \item We define the fundamental JEPA mechanism for wireless systems, including context and target views, online and target encoders, momentum target updates, latent prediction loss, and collapse-avoidance intuition.
    \item We present a wireless-oriented tokenization and masking framework that maps CSI, beam-domain observations, KPIs, topology graphs, and ISAC measurements into a unified JEPA-compatible token sequence.
    \item We clarify the three operational phases of wireless JEPA, such as self-supervised pretraining, task adaptation with lightweight heads or controllers, and online deployment under RAN, edge, core, and O-RAN constraints.
    \item We provide compact, comparable JEPA design recipes for representative 6G use cases, including beam management, link adaptation, ISAC, traffic load, O-RAN automation, and security, each specifying context tokens, latent target, task head, label requirement, and deployment location.
    \item We provide an illustrative, synthetic beam-management case study using a beam-aware Future-JEPA instantiation, which isolates the effect of wireless-aware target design and suggests that an auxiliary future beam-energy target during self-supervised pretraining can improve label efficiency and robustness across shifted deployment conditions relative to a supervised source domain.
    \item We identify open challenges in long-horizon prediction, action-conditioned modeling, distributed training, trustworthiness, efficiency, benchmarking, and standardization.
\end{itemize}

\section{JEPA Fundamentals}

\subsection{Core Principles of JEPA for 6G}

JEPA is an SSL paradigm that learns predictive representations in latent space. Given a wireless observation window, JEPA forms two related views: a context view containing visible or currently available information, and a target view containing masked, missing, or future information to be predicted. Unlike reconstruction-based SSL, JEPA does not recover raw target samples. Instead, it predicts the target embedding produced by a separate target encoder, which encourages the model to capture abstract, stable, and decision-relevant structure rather than low-level measurement details \cite{chu2026wirelessjepa}.

This principle can support 6G because wireless and network observations are noisy, incomplete, distributed, and affected by channel variation, interference, mobility, blockage, and hardware impairments. Many 6G tasks require inferring a predictive network state rather than classifying a single snapshot, such as anticipating beam quality, estimating link reliability, predicting congestion, detecting anomalous behavior, and inferring environment structure from ISAC measurements.

\begin{figure}[t]
\centering
\includegraphics[width=\columnwidth]{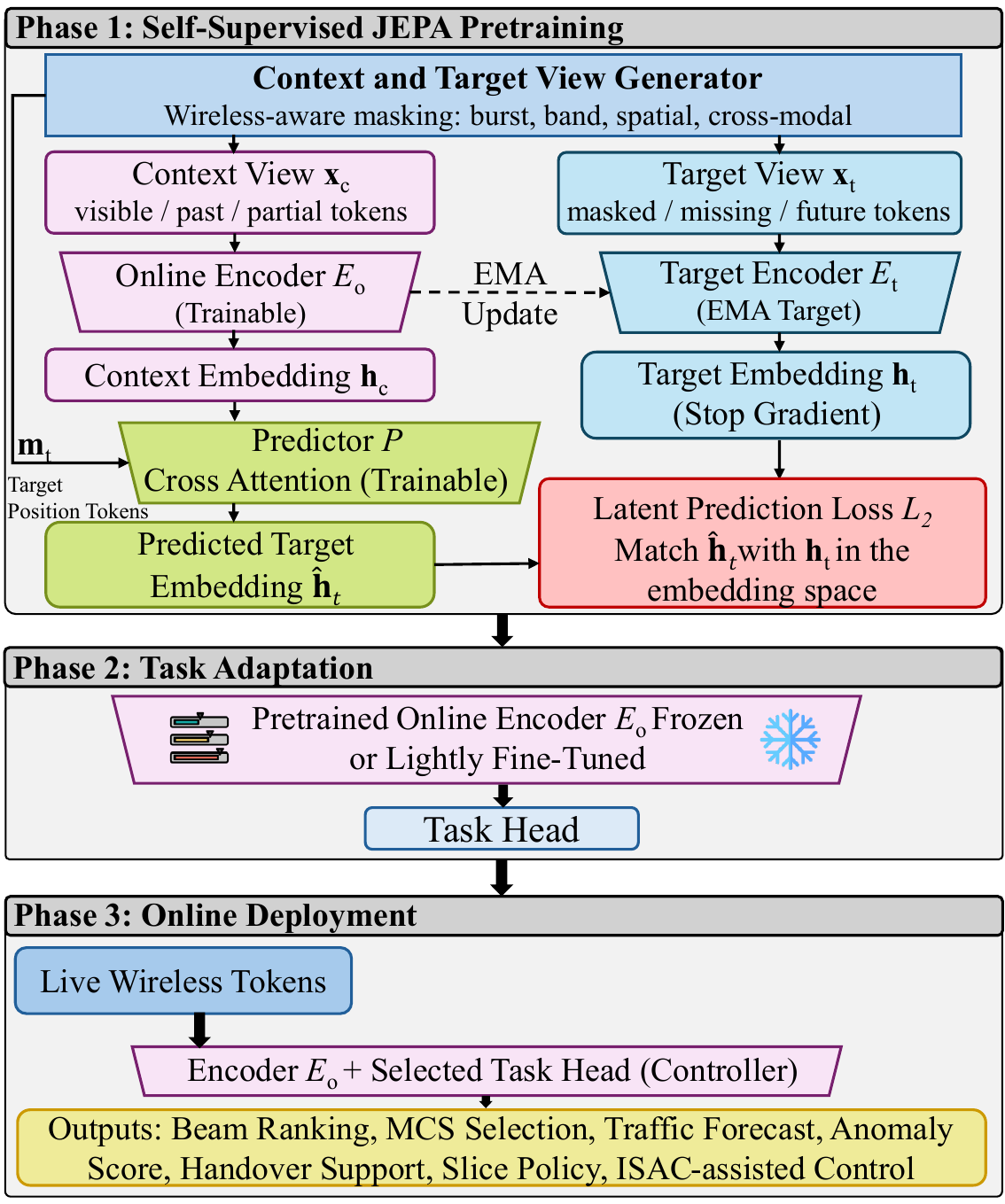}
\caption{Three-phase wireless JEPA workflow. \textbf{Phase 1 (pretraining):}
the online encoder $E_o$ and predictor $P$ are trained to match stop-gradient target embeddings from the target encoder $E_t$, which is updated by EMA; optional target-position tokens $\mathbf{m}_t$ select predicted locations. \textbf{Phase 2 (adaptation):} $E_o$ is frozen or lightly fine-tuned and a lightweight task-specific head/controller is trained. \textbf{Phase 3 (deployment):} $E_o$ plus the selected head produce beam ranking, MCS selection, traffic forecasts, anomaly scores, handover support, slice policy, or ISAC-assisted control. }
\label{fig:jepa_framework}
\end{figure}

\subsection{JEPA Architecture}

A JEPA model contains three main components: an online encoder, a target encoder, and a predictor, as illustrated in Fig.~\ref{fig:jepa_framework}. The online encoder $E_o(\cdot;\theta_o)$ processes the context view $\mathbf{x}_c$ and produces a context embedding. The target encoder $E_t(\cdot;\theta_t)$ processes the target view $\mathbf{x}_t$ and produces the target embedding. The context and target views $\mathbf{x}_c$ and $\mathbf{x}_t$ are drawn from the unified token sequence constructed in Fig.~\ref{fig:jepa_inputs}. The predictor $P(\cdot;\phi)$ maps the context embedding toward the target embedding, optionally conditioned on target-position tokens $\mathbf{m}_t$ that indicate which time, frequency, beam, or modality positions are being predicted, so that a single predictor can query multiple target locations. However, in the beam-management case study given in Sec.~\ref{sec:case_study}, the predictor queries a single pooled future target, so $\mathbf{m}_t$ is not used. During pretraining, gradients update the online encoder and predictor, while the target encoder is updated by an exponential moving average (EMA) of the online encoder parameters. This momentum update provides slowly evolving target representations and improves training stability \cite{Assran_2023_CVPR}.

In 6G wireless systems, the context view may contain visible CSI tiles, recent beam measurements, KPI windows, topology tokens, or sensing tokens. The target view may correspond to masked tokens from the same time window or future tokens from a later window. For example, in beam management, the context can be a history of partial beam-power scans, while the target can be a future beam-domain representation. In O-RAN automation, the context can be recent KPI and topology tokens, while the target can be a short-horizon network-state embedding. Moreover, JEPA is separated from the downstream decision module. It outputs embeddings, not beams, MCS values, handover actions, slice policies, or anomaly labels. After pretraining, the online encoder is reused as a representation module, and a lightweight task head or controller is trained for the required function. The target encoder and predictor are primarily training-side modules, although prediction residuals can also be used for uncertainty or anomaly scoring.

\subsection{Training Objectives \& Collapse Avoidance}

The JEPA objective aligns a representation inferred from the observed wireless context with the representation of a masked or future target. The online encoder maps the context into $\mathbf{h}_c = E_o(\mathbf{x}_c;\theta_o)$, while the target encoder maps the target view into the \emph{latent} target embedding $\mathbf{h}_t = E_t(\mathbf{x}_t;\theta_t)$. Then, the predictor produces $\hat{\mathbf{h}}_t = P(\mathbf{h}_c;\phi)$, which is trained to match the stop-gradient target embedding $\mathrm{sg}(\mathbf{h}_t)$. Thus, gradient descent updates $\theta_o$ and $\phi$, while the target parameters $\theta_t$ are updated by EMA, i.e., $\theta_t \leftarrow \tau\theta_t + (1-\tau)\theta_o$, where $\tau \in [0,1)$ is the momentum coefficient. Depending on the implementation, the loss can be computed at the token level, block level, or after pooling the target representation. Moreover, the beam-management case study given in Sec.~\ref{sec:case_study} uses a pooled context embedding to match a pooled future target. We emphasize that $\mathbf{h}_t$ is a \emph{learned latent} target produced by the target encoder; this is distinct from any auxiliary, hand-designed target defined directly in the measurement domain (e.g., the future beam-energy distribution used in the case study in Sec.~\ref{sec:case_study}), which we introduce separately to keep the two notions clearly apart.

Because the loss is computed in embedding space rather than raw input space, JEPA is not required to reconstruct noisy I/Q samples, detailed CSI tensors, or low-level KPI fluctuations. Instead, it learns to preserve information predictive of the target representation, which is useful when measurements include thermal noise, pilot contamination, quantization effects, hardware artifacts, and task-irrelevant variations. A major concern in non-contrastive SSL is representation collapse, where the model produces uninformative constant embeddings \cite{li2022understanding}. JEPA mitigates this risk through different contexts and target views, predictor asymmetry, and slowly updated EMA targets. In 6G wireless systems, additional normalization or variance regularization can also be useful because neighboring time-frequency tiles, adjacent beams, and consecutive KPI samples are often highly correlated. Without careful masking, target design, and regularization, an SSL model may learn shortcuts rather than meaningful radio-environment structure.

\subsection{JEPA Comparison with Contrastive \& Reconstruction SSL}

JEPA complements reconstruction-based and contrastive SSL rather than replacing them. Reconstruction-based models recover corrupted or masked inputs and can be effective when the target is carefully selected, e.g., structured channel features rather than raw noisy samples. However, if the reconstruction target is too close to the raw measurement domain, the model may spend capacity on noise, hardware artifacts, or fine-grained fluctuations that are not essential for control. Contrastive methods learn by pulling positive pairs together and pushing negative pairs apart. Their success depends on wireless-preserving augmentations and meaningful negative samples, which are task- and channel-dependent. For example, time shifts, frequency masking, phase perturbations, or antenna-domain transformations may be valid for some tasks but harmful for beam prediction, channel estimation, or localization.

Therefore, JEPA avoids explicit raw-sample reconstruction and negative-sample construction by predicting target representations from context representations. The key design question is not whether JEPA is universally better than other SSL methods, but whether the latent target is aligned with the downstream wireless decision. As shown in the beam-management case study in Fig.~\ref{fig:jepa_case_study}, JEPA is most useful when the predictive target preserves future control-relevant structure, such as the future beam-energy distribution. Table~\ref{tab:ssl_comparison} summarizes this comparison, and presents it as a design tradeoff rather than a ranking: no single paradigm dominates universally, each requires a different target or augmentation design, and JEPA's main risk is a poor latent-target choice.

\begin{table}[t]
\centering
\scriptsize
\caption{Comparison of SSL paradigms for wireless representation learning.}
\label{tab:ssl_comparison}
\renewcommand{\arraystretch}{1.05}
\begin{tabularx}{\columnwidth}{
>{\raggedright\arraybackslash}p{0.16\columnwidth}
>{\raggedright\arraybackslash}p{0.20\columnwidth}
>{\raggedright\arraybackslash}p{0.20\columnwidth}
>{\raggedright\arraybackslash}X}
\toprule
\textbf{SSL type} & \textbf{Objective} & \textbf{Wireless challenge} & \textbf{Main risk} \\
\midrule
Reconstruction / MAE &
Recover masked raw or structured inputs &
Raw CSI or I/Q may contain noise and hardware artifacts &
Overfitting low-level details \\
Contrastive SSL &
Pull positives together and push negatives apart &
Wireless-preserving augmentations are task-dependent &
Wrong invariances may remove useful information \\
JEPA &
Predict target embeddings from context embeddings &
Target must match the wireless control objective &
Poor target design may reduce downstream gain \\
\bottomrule
\end{tabularx}
\end{table}

\section{From Wireless Data to JEPA Tokens \& Masking}
\label{sec:tokens}

\begin{figure}[t]
\centering
\includegraphics[width=\columnwidth]{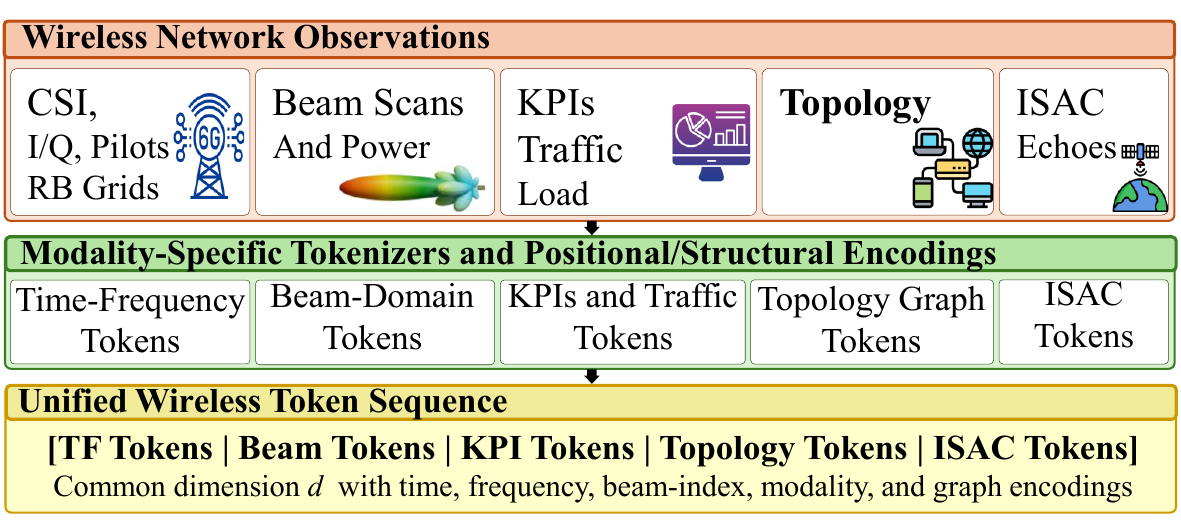}
\caption{From heterogeneous wireless and network observations to a unified
JEPA token sequence. Modality-specific tokenizers map CSI/I-Q, beam scans,
KPIs/traffic, topology, and ISAC observations into a common $d$-dimensional
token space with positional, temporal, frequency, beam-index, modality, and
graph-structural encodings. Variable-size modalities (e.g., topology graphs) are reduced to a stable token count via pooling or structural summaries, and missing or asynchronous streams are handled with padding masks.}
\label{fig:jepa_inputs}
\end{figure}

A practical wireless JEPA model starts by converting heterogeneous wireless and network observations into a common token representation suitable for latent prediction. Unlike language or vision, where inputs have a relatively uniform structure, 6G data span multiple physical and logical layers, including I/Q samples, OFDM resource grids, CSI, beam measurements, RAN KPIs, topology graphs, mobility traces, and ISAC observations. These modalities differ in sampling rate, dimension, reliability, and semantics \cite{masserano2024enhancing}. Therefore, a wireless-oriented JEPA framework must preserve task-relevant structure while supporting missing, asynchronous, and variable-size inputs. The tokenization pipeline is summarized in Fig.~\ref{fig:jepa_inputs}, and the resulting sequence feeds the context and target view generator of Fig.~\ref{fig:jepa_framework}.

\subsection{Tokenization of Wireless Network Data}

Let an observation window contain modality set $ \mathcal{M}=\{\mathrm{tf},b,k,g,s\} $, representing time--frequency, beam-domain, KPI, topology, and sensing information, respectively. Each modality is processed by a modality-specific tokenizer and projected to a common embedding dimension $ d $. The token stream of modality $ m $ is denoted by $ \mathbf{X}_m \in \mathbb{R}^{N_m \times d} $, where $ N_m $ is the number of tokens for that modality. Since $ N_m $ may vary across cells, users, beams, sensing snapshots, and topology nodes, padding masks, pooling, local neighborhoods, or graph-to-token conversion can be used to handle variable-size inputs. Positional, temporal, frequency, beam-index, modality, and graph-structural encodings are added to help the JEPA encoder distinguish tokens' origins and structures.

\begin{itemize}
    \item \textit{Time-frequency tokens} are obtained by partitioning the OFDM resource grid, CSI tensor, or pilot observations into local time-frequency tiles. They capture channel coherence, interference patterns, Doppler variation, and scheduling-relevant structure \cite{guler2026pilotwimae}.
    
    \item \textit{Beam-domain tokens} are derived from antenna-array responses, beam sweeping measurements, beam-power scans, and beam quality reports. They represent directional propagation structure and are useful for mmWave and sub-THz beam management.
    
    \item \textit{KPI and traffic tokens} encode indicators such as SINR, RSRP, throughput, latency, cell load, packet loss, handover statistics, and slice-level telemetry over sliding windows. They support traffic forecasting, anomaly detection, and O-RAN automation.
    
    \item \textit{Topology tokens} represent the network as a graph of UEs, gNBs, edge servers, and core-network functions. Node and edge features can be converted into graph tokens using graph neural encoders, neighborhood pooling, or structural embeddings; the number of graph tokens varies with the active topology, so pooling or fixed-size structural summaries are used to obtain a stable token count.
    
    \item \textit{Sensing tokens} encode ISAC observations such as radar-like echoes, localization cues, blockage indicators, and environmental measurements. They provide context for beam alignment, blockage prediction, localization-aware scheduling, and environment-aware control.
\end{itemize}

After tokenization, the modality-specific streams are concatenated along the token axis to form a unified sequence $\mathbf{Z}=[\mathbf{X}_{\mathrm{tf}};\mathbf{X}_{b};\mathbf{X}_{k};\mathbf{X}_{g};\mathbf{X}_{s}] \in \mathbb{R}^{N \times d}$, where $N=\sum_{m\in\mathcal{M}}N_m$. This concatenation does not directly combine raw measurements; each modality is first projected to the shared $d$-dimensional token space. We stress that this token-level concatenation is not a trivial merge, as modalities arrive at different rates and reference times, so a common temporal grid, resampling or interpolation, and explicit modality/time encodings are required to align tokens, and missing or asynchronous streams are represented through padding masks and cross-modal masking rather than silently dropped. Then the unified sequence $\mathbf{Z}$ is processed by the JEPA view generator, which selects visible tokens to form the context view $\mathbf{x}_c$ and masked or future tokens to define the target view $\mathbf{x}_t$; when the predictor is queried at specific locations, the corresponding target-position tokens $\mathbf{m}_t$ index the predicted time, frequency, beam, or modality positions.

A wireless JEPA training step can be summarized in three stages. First, modality-specific tokenizers map raw observations into common $d$-dimensional tokens and concatenate them into $\mathbf{Z}$. Second, a wireless-aware mask or future-view operator forms the context view $\mathbf{x}_c$ and target view $\mathbf{x}_t$. Third, the online encoder and predictor are trained to match the target embedding, while the target encoder is updated by EMA. After pretraining, the online encoder is reused with lightweight task heads or controllers.

\subsection{Masking Strategies for Wireless Impairments}

The masking strategy defines the self-supervised prediction task. In wireless systems, masks should not only regularize training but also reflect realistic sources of partial observability, measurement loss, and domain shift \cite{guler2025multitask}. A useful masking policy determines what the online encoder observes, what the target encoder represents, and what predictive structure the model is forced to learn.

\begin{itemize}
    \item \textit{Burst masking} removes contiguous time-domain regions, emulating packet losses, handover gaps, fading bursts, intermittent sensing availability, and temporary telemetry interruptions.
    
    \item \textit{Band masking} removes frequency sub-bands, reflecting narrowband interference, frequency-selective fading, carrier aggregation gaps, spectrum sharing, and missing pilot resources.
    
    \item \textit{Spatial masking} removes beam, antenna-region, or angular-domain tokens, encouraging the model to infer spatial channel structure from partial observations. This is useful for blockage-aware beam management and beam recovery.
    
    \item \textit{Cross-modal masking} removes an entire modality, such as sensing, KPI, topology, CSI, or beam tokens. This represents practical scenarios in which a data stream is delayed, corrupted, unavailable, asynchronous, or unsupported by a specific network segment.
    
    \item \textit{Future masking} uses a current or past context window to predict tokens from a later horizon. This is important for proactive control tasks such as future beam prediction, link-risk estimation, congestion anticipation, and mobility-aware resource management.
\end{itemize}

These masking policies expose the JEPA encoder to realistic missing data and partial observability. The mask choice should match the downstream task, such as beam management benefits from spatial and future masking, traffic forecasting from burst and future KPI masking, and ISAC from cross-modal masking between sensing and communication tokens.

\section{JEPA Integration in 6G Architectures}

As illustrated in Fig.~\ref{fig:jepa_framework}, JEPA can be integrated into 6G networks as a shared predictive representation layer across the RAN, edge, and core. The JEPA encoder transforms heterogeneous wireless and network observations into compact embeddings that support control, optimization, and analytics tasks. Therefore, JEPA should be viewed as a representation module rather than a controller: it predicts latent network states, while task-specific heads or classical controllers produce beams, MCS selections, handover actions, slice policies, or anomaly scores.

This separation leads to three operational phases. First, during self-supervised pretraining, the online and target encoders are trained using masked or future wireless targets. Second, during task adaptation, the pretrained online encoder is frozen or lightly fine-tuned and connected to lightweight heads trained with limited labels, or used for latent-residual scoring. Third, during online deployment, only the online encoder and the selected task head are executed inside the live control loop. This design is relevant for 6G because future control loops will be multimodal, partially observable, non-stationary across terrestrial and non-terrestrial segments, and constrained by O-RAN latency and multi-vendor lifecycle requirements.

\subsection{JEPA as a Shared Predictive Encoder for RAN and Core Functions}

In AI-native 6G systems, intelligence is expected to be distributed across users, RAN nodes, edge platforms, and core-network functions. This motivates a shared encoder that exposes predictive features to multiple functions instead of training a separate model for each task. JEPA can act as such a shared predictive encoder, following the latent-prediction principle established in I-JEPA \cite{Assran_2023_CVPR} and extended to multi-antenna wireless I/Q data in WirelessJEPA \cite{chu2026wirelessjepa}. Reaching a true foundation-model scale would additionally require large-scale multi-site pretraining data and systematic transfer evaluation, which remain open for wireless. It can convert CSI, beam measurements, HARQ feedback, traffic patterns, mobility traces, and sensing observations into reusable predictive embeddings.

At the RAN level, a frozen or fine-tuned online encoder can feed lightweight heads for beam management, link adaptation, scheduling, interference coordination, and mobility control. At the core-network level, the same embeddings can support traffic forecasting, slice admission control, service-level agreement (SLA) risk prediction, anomaly detection, and policy optimization. For anomaly detection, latent prediction residuals or embedding-density scores can be useful even when labeled attack data are limited.

The deployment location should match the data source and latency requirement. UE and gNB-based encoders can consume CSI, I/Q samples, and beam reports under tight latency and privacy constraints. Edge or near-RT RIC instances can fuse cell-level KPIs and topology information, while core or non-RT functions can aggregate traffic and SLA telemetry. Thus, a monolithic encoder assuming all modalities are centrally available is different from a sparse or distilled encoder designed for practical gNB or near-RT RIC deployment.

\subsection{Multi-Modal and Cross-Layer Embedding Fusion}

6G networks must process wireless and network modalities such as RF signals, sensing data, topology information, traffic telemetry, and SLA requirements, where \emph{multimodal} refers to these network and wireless modalities rather than only camera or lidar sensor streams. JEPA fuses these by mapping each modality through a dedicated tokenizer into a shared $d$-dimensional embedding space. The resulting tokens are concatenated along the sequence axis and processed jointly, so that fusion occurs through latent-space attention rather than direct addition of raw signals.

This design enables cross-modal prediction under missing or asynchronous observations. For example, CSI and mobility tokens can support proactive scheduling, while sensing and beam-domain tokens can improve blockage prediction and beam alignment. Since modalities arrive at different rates and are intermittently unavailable, masked latent prediction allows the encoder to infer the latent content of an absent or low-rate modality from available observations. Cross-layer fusion is also important because conventional protocol stacks maintain separate PHY, MAC, network, and application-layer control mechanisms \cite{10504601}. A JEPA encoder exposes shared predictive features across these layers, such as PHY channel trends that can guide MAC scheduling, QoS targets that can shape link adaptation, and mobility or sensing context that can support handover and routing.

\subsection{O-RAN-Compliant Integration and Interfaces}

The O-RAN framework provides a practical integration path for JEPA through near-RT RIC xApps and non-RT RIC rApps \cite{10024837}. A large JEPA model can be pretrained or periodically updated in the non-RT RIC using historical data collected through O1 interfaces and, where relevant, O2 cloud or infrastructure interfaces. Then the non-RT RIC can provide policy guidance, enrichment information, and model-related recommendations to the near-RT RIC through A1. A distilled online encoder with its task-specific head can run as a near-RT xApp, using E2 telemetry from E2 nodes and sending control recommendations through the E2 interface \cite{10024837}.

Latency plays an important role in this integration, where near-RT RIC control typically operates on the order of $10\,\mathrm{ms}$--$1\,\mathrm{s}$, while non-RT RIC functions operate above $1\,\mathrm{s}$. Moreover, robust PHY and MAC decisions require gNB-local or DU-local deployment rather than near-RT RIC inference. Therefore, JEPA-enabled xApps for beam prediction, scheduling, or link adaptation require lightweight encoders, efficient tokenization, sparse or pooled token sets, model compression, and possibly edge hardware acceleration.

Standard O-RAN interfaces can provide telemetry such as UE measurements, CSI-related statistics, traffic counters, cell-load indicators, and topology information. However, practical deployment also requires embedding exchange, model lifecycle management, privacy protection, and interoperability across vendors. Exchanging embeddings rather than raw data can reduce bandwidth and privacy exposure, but only if network entities agree on a versioned latent space. Therefore, embedding schemas, drift monitoring, model rollback, and compatibility testing should be part of the O-RAN AI lifecycle.

\subsection{Operational Considerations and Data Governance}

JEPA deployment in operational 6G networks raises privacy, governance, and reliability issues. Network data reveal sensitive information about user behavior, service usage, mobility patterns, and infrastructure vulnerabilities. Privacy-preserving and federated JEPA training reduces the need to centralize raw data while enabling collaborative model improvement. In a federated setting, clients may share encoder updates, target-encoder momentum updates, or embeddings rather than raw CSI or traffic traces, optionally protected by secure aggregation or differential privacy \cite{10944600}.

JEPA models must also remain reliable under distribution shifts, missing modalities, and adversarial manipulation. Therefore, uncertainty estimation, anomaly scoring, and out-of-distribution (OOD) detection should be integrated into JEPA-based control loops. Latent prediction residuals can indicate when an embedding has drifted far enough that the downstream controller should fall back to a conservative baseline. Before field deployment, JEPA-based control should be validated through realistic testbeds, digital twins (DTs), and hardware-in-the-loop (HIL) experiments \cite{10944600}. O-RAN testbeds, Colosseum-like emulation platforms, and network DTs can evaluate JEPA under realistic wireless dynamics, multi-vendor constraints, and operational lifecycle requirements.

\begin{table*}[t]
\scriptsize
\centering
\caption{Comparable JEPA design recipes for representative 6G use cases, each specifying context tokens, latent prediction target, task head/output, label requirement, and deployment location.}
\label{tab:usecase_map}
\renewcommand{\arraystretch}{1.15}
\setlength{\tabcolsep}{4pt}
\begin{tabularx}{\textwidth}{
>{\raggedright\arraybackslash}p{0.11\textwidth}
>{\raggedright\arraybackslash}p{0.19\textwidth}
>{\raggedright\arraybackslash}p{0.17\textwidth}
>{\raggedright\arraybackslash}p{0.18\textwidth}
>{\raggedright\arraybackslash}p{0.13\textwidth}
>{\raggedright\arraybackslash}X}
\toprule
\textbf{Use case} & \textbf{Context tokens} & \textbf{Latent target} & \textbf{Task head/output} & \textbf{Label requirement} & \textbf{Deployment} \\
\midrule
\textbf{Beam management} &
CSI tiles, beam history, UE motion, topology &
Future or masked beam-domain embedding &
Beam ranking, handover support, blockage-aware control &
Small best-beam/top-$k$ set for probe &
gNB/DU-local or near-RT RIC \\
\midrule
\textbf{Link adaptation} &
CSI, SINR/CQI, HARQ, interference, mobility &
Future link-state embedding (reliability/BLER trend) &
Risk-aware MCS, margin/rate selection &
Few ACK/NACK or BLER labels for head &
gNB/DU-local (fast loop) \\
\midrule
\textbf{ISAC} &
Partial CSI, radar-like echoes, localization cues &
Masked sensing / future channel tokens; environment embedding &
Proactive beam alignment, environment-aware scheduling, positioning &
Sensing self-supervised; few positioning/blockage labels &
gNB or co-located edge \\
\midrule
\textbf{Traffic \& O-RAN} &
RAN KPI windows, load, topology graph, slice telemetry &
Masked or future network-state embedding (short horizon) &
Load forecast, congestion/admission control, slice orchestration &
Targets self-generated from future KPIs &
near-RT RIC (short) / non-RT RIC (slow) \\
\midrule
\textbf{Security \& resilience} &
I/Q, KPIs, topology, telemetry (normal state) &
Normal-state embedding; anomaly via latent residuals &
Jamming/spoofing detection, resilient fallback &
Label-free normal pretraining; optional few attack labels &
gNB/edge detection + non-RT monitoring \\
\bottomrule
\end{tabularx}
\end{table*}

\section{Representative Use Cases}
\label{sec:usecases}

JEPA applies to 6G scenarios where the network must infer missing or future states from partial, noisy, and multimodal observations. It does not replace conventional communication algorithms; rather, it provides predictive embeddings that support downstream control, optimization, and decision-making. Table~\ref{tab:usecase_map} summarizes five representative use cases as comparable design methods, each specifying the context tokens, latent prediction target, task head/output, label requirement, and deployment location.

\subsection{Beam Management and Blockage-Aware Mobility}

Beam management in mmWave and sub-THz systems is a representative JEPA use case because link quality depends strongly on directional alignment and is sensitive to blockage, mobility, and environmental changes. Conventional supervised beam prediction requires labeled beam indices from beam sweeping or exhaustive measurements and can degrade under domain shifts across propagation conditions, hardware platforms, or deployment sites. Instead, a JEPA-based pipeline learns a predictive representation of the evolving radio environment from partially observed context.

\textit{Pretraining phase:} An observation window includes recent CSI tiles, beam-history measurements, short-horizon mobility traces, and KPIs related to link quality or handover status. These inputs are converted into time-frequency, beam-domain, and KPI tokens and concatenated into a unified sequence. During JEPA pretraining, the online encoder receives visible context tokens, while the target encoder processes selected masked or future target tokens used only as stop-gradient latent targets. The predictor learns to infer target embeddings from context embeddings, allowing the encoder to capture directional structure, blockage evolution, and mobility-induced channel changes without reconstructing raw measurements.

\textit{Task-adaptation phase:} After pretraining, the online encoder is reused as a feature extractor. A lightweight beam-ranking head is trained using a limited labeled dataset, where labels represent the best beam, top-$k$ beam list, beam-failure risk, or handover trigger. Since the encoder has already learned predictive structure from unlabeled observations, adaptation requires fewer labels than training from scratch.

\textit{Deployment phase:} During online operation, the current observation window is encoded into a predictive radio-environment embedding. A downstream beam-management head uses this embedding for beam ranking, beam-failure recovery, proactive handover preparation, or blockage-aware beam switching. Because beam decisions are latency-critical, this encoder--head pair is best placed in the gNB or, for slower coordination, as a near-RT RIC xApp. The same representation can also support related tasks such as blockage prediction, mobility-aware beam adaptation, and cross-cell coordination.

\subsection{Robust Link Adaptation and Rate Selection}

Link adaptation selects modulation, coding, and transmission parameters under uncertain channel and interference conditions. Conventional methods rely on SINR, CQI, and HARQ feedback, which are useful but only partially describe the link state. \textit{Context:} recent CSI, SINR/CQI reports, HARQ outcomes, interference indicators, and mobility tokens over a sliding window. \textit{Latent target:} the embedding of a short future window, so that the encoder learns to predict reliability trends such as impending BLER increase or outage risk. \textit{Head:} a lightweight regression or classification head mapping the embedding to a risk-aware MCS, transmission margin, or reliability-aware schedule. \textit{Labels:} the head can be supervised with naturally generated operational feedback such as per-transmission ACK/NACK outcomes, or with block-error-rate (BLER) estimates aggregated over multiple transmissions; only a small number are needed, since the reliability structure is already captured during self-supervised pretraining. \textit{Deployment:} because rate selection operates on a fast loop, the encoder--head pair should run gNB rather than in the near-RT RIC.

\subsection{Integrated Sensing and Communications}

ISAC systems generate correlated communication and sensing observations, including CSI, radar-like echoes, localization cues, and environmental context. Instead of processing sensing and communication features separately, JEPA can learn a shared latent space across modalities. \textit{Context:} partial CSI together with recent sensing observations (radar-like echoes, localization cues). \textit{Latent target:} masked sensing tokens or future channel tokens, so the predictor is forced to infer an environment/blockage-aware cross-modal embedding rather than reconstruct raw echoes. \textit{Head:} task heads for proactive beam alignment, environment-aware scheduling, joint communication--sensing resource allocation, or a positioning regressor. \textit{Labels:} the sensing-to-communication prediction is self-supervised; only a few positioning or blockage labels are needed to calibrate the corresponding head. \textit{Deployment:} at the gNB or a co-located edge node where sensing and communication data are jointly available \cite{3gpp_tr38843}.

\subsection{Traffic Prediction and O-RAN Closed-Loop Automation}

O-RAN automation depends on predicted traffic, mobility, congestion, and slice-level demand. Existing xApps and rApps rely on task-specific forecasting models trained for individual KPIs or cells, which may have limited transferability. \textit{Context:} sliding windows of RAN KPIs, load indicators, topology graph tokens, mobility traces, and slice telemetry. \textit{Latent target:} a masked or future network-state embedding at a chosen prediction horizon, capturing short-horizon congestion and demand evolution. \textit{Head:} lightweight forecasting and control heads for load prediction, congestion anticipation, mobility-aware load balancing, admission control, energy-saving cell activation, and slice orchestration. \textit{Labels:} prediction targets are largely self-generated from future KPI values, so little manual labeling is needed. \textit{Deployment:} short-horizon forecasts fit near-RT RIC xApps, while slower slice- and energy-management decisions map to non-RT RIC rApps \cite{10024837}.

\subsection{Security, Anomaly Detection, and Resilience}

Security and resilience are critical for 6G networks operating in dynamic or adversarial environments. Relevant threats and faults include jamming, spoofing, telemetry corruption, sensing manipulation, control-plane instability, and distributed infrastructure failures. Since labeled attack data are often scarce, the natural JEPA recipe is anomaly detection by \emph{normal-state} modeling rather than supervised attack classification. This assumes the pretraining traces are predominantly clean; if the unlabeled data may itself be poisoned or adversarially corrupted, robust or filtered pretraining is required, which is a distinct problem beyond the scope of this recipe. \textit{Context:} I/Q streams, KPIs, topology tokens, and telemetry collected during normal (or mostly normal) operation. \textit{Latent target:} the predictive embedding of the normal network state. \textit{Head:} a residual/embedding-density scorer that flags anomalies when the observed target embedding deviates from the predicted one; where a limited set of labeled attacks is available, a small supervised classifier can be added as a second mode. \textit{Labels:} the primary mode is label-free (normal-state pretraining with residual scoring); only the optional supervised mode requires a few attack labels. \textit{Deployment:} fast detection at the gNB/edge, with aggregated monitoring and trust-aware fallback coordinated by non-RT functions.

\section{Case Study}
\label{sec:case_study}

\begin{table}[t]
\scriptsize
\centering
\caption{Beam-management case-study setup. A domain is a tuple (SNR~dB, blockage prob., angular speed deg./step, beam-keep prob., time-drop prob., sector-drop prob.).}
\label{tab:case_config}
\renewcommand{\arraystretch}{1.1}
\setlength{\tabcolsep}{4pt}
\begin{tabularx}{\columnwidth}{
>{\raggedright\arraybackslash}p{0.24\columnwidth}
>{\raggedright\arraybackslash}X}
\toprule
\textbf{Item} & \textbf{Value} \\
\midrule
Array / codebook & $32$-antenna ULA, $32$-beam codebook. \\
Channel model & Clustered geometric, $3$--$6$ paths, angular drift, random dominant-path blockage, noisy log beam-power scans with partial observation. \\
History/horizon & Observe $8$ noisy scans; predict best beam $3$ steps ahead (SSL target: future $3$ normalized scans). \\
Data (disjoint) & $30{,}000$ unlabeled SSL, $10{,}000$ labeled pool, $3{,}000$ test per domain; independently generated trajectories. \\
Train domain (SSL) & Randomized: SNR $0$--$25$, blockage $0.02$--$0.35$, speed $0.5$--$4.5$. \\
Source / ID domain & $(20,0.05,1.5,0.75,0.05,0.10)$. \\
Combined OOD & $(5,0.30,3.8,0.45,0.20,0.35)$. \\
Isolated OOD & Low-SNR $(0,\!\cdot)$; blockage $(\cdot,0.40,\!\cdot)$; mobility $(\cdot,5.0,\!\cdot)$; missing $(\cdot,0.35,0.25,0.45)$. \\
Labels & Frozen-encoder probe at $1\%,5\%,10\%$ source labels ($100/500/1{,}000$ trajectories). \\
Baselines & SSL: BA-Future-JEPA, Future-JEPA, MAE, SimCLR; supervised scratch; non-learned: last-best, history-mean, random. \\
Metrics & Top-1/Top-3 accuracy, normalized beamforming gain (linear), refined gain rGain@$K$, coverage Cov@90. \\
\bottomrule
\end{tabularx}

\vspace{2pt}
{\scriptsize \textit{Fixed a-priori hyperparameters (not tuned on any test domain):} Transformer encoder $d{=}96$, $4$ heads, $3$ layers, dropout $0.10$; AdamW, lr $3{\times}10^{-4}$, weight decay $10^{-4}$, $50$ SSL epochs, batch $256$; EMA $\tau{=}0.996$; variance-regularization weight $0.02$; MAE temporal-block mask ratio $0.35$; BA-Future-JEPA beam temperature $0.50$, beam-target weight $1.00$; probe/scratch $60$ epochs; $5$ seeds.}
\end{table}

\begin{figure*}[t]
    \centering
    \includegraphics[width=\textwidth]{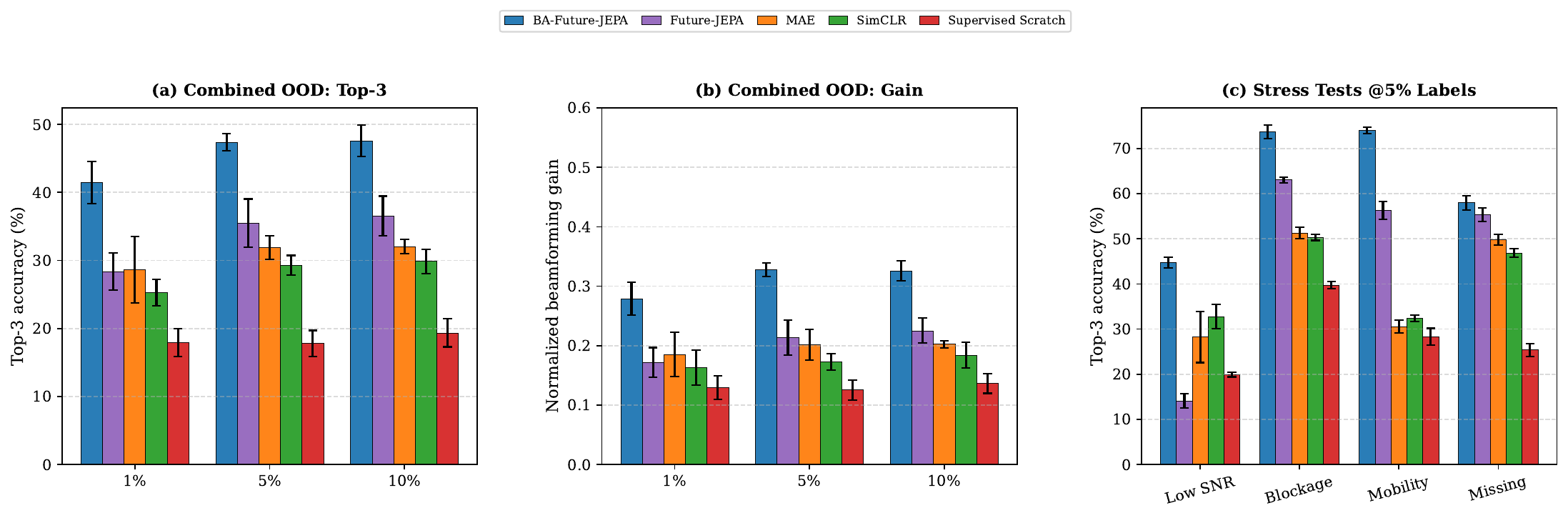}
    \caption{Label-efficiency and robustness of BA-Future-JEPA compared with generic Future-JEPA, MAE, SimCLR, and supervised scratch (mean over five seeds). (a)--(b) Combined OOD Top-3 accuracy and normalized beamforming gain across label fractions; (c) Top-3 accuracy at $5\%$ labels under isolated OOD impairments. }
    \label{fig:jepa_case_study}
\end{figure*}

\begin{figure*}[t]
    \centering
    \includegraphics[width=\textwidth]{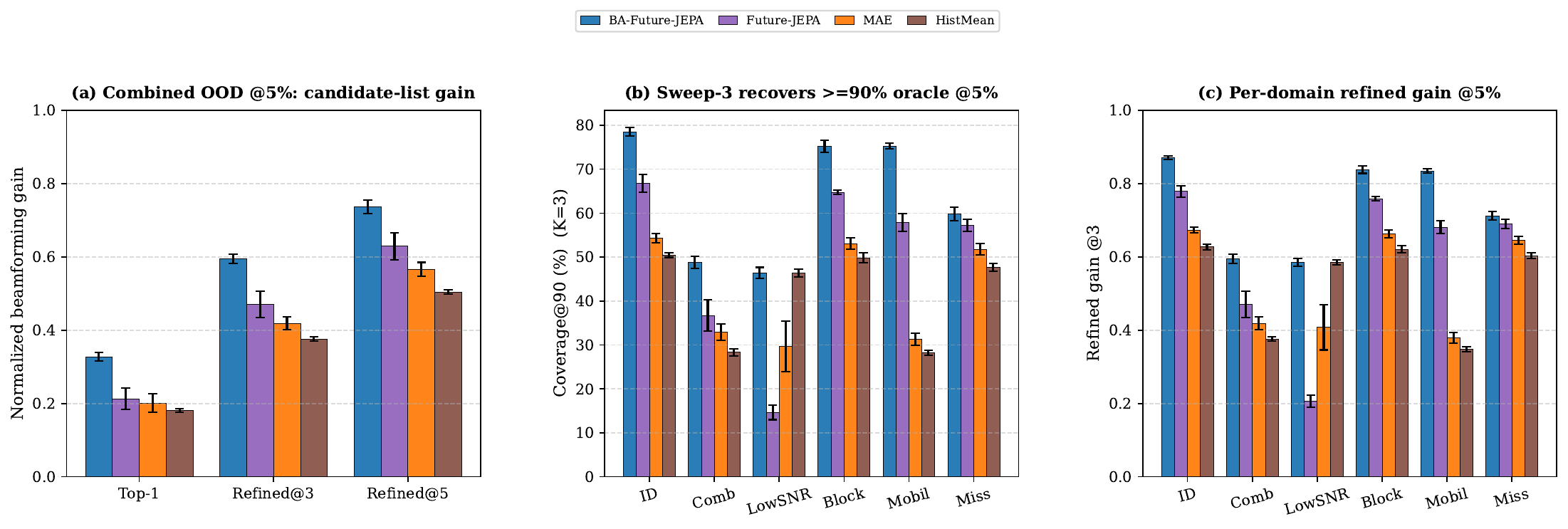}
    \caption{Operational candidate-list view, comparing BA-Future-JEPA with generic Future-JEPA, MAE, and a non-learned persistence baseline (history-mean). (a) Refined gain rGain@$K$ on combined OOD at $5\%$ labels; (b) per-domain Cov@90 for Top-3 candidate sweeping; (c) per-domain rGain@3. }
    \label{fig:jepa_refined}
\end{figure*}

\begin{table*}[t]
\scriptsize
\centering
\caption{Beam-management results on the combined worst-case OOD domain at $5\%$ source labels (mean $\pm$ sample std over five seeds). Gain is the linear normalized beamforming gain of the selected beam; rGain@$K$ is the gain when the true best of the predicted Top-$K$ candidates is used; Cov@90 is the fraction of cases where sweeping the Top-3 candidates recovers $\ge 90\%$ of oracle beam power.}
\label{tab:case_results}
\renewcommand{\arraystretch}{1.15}
\setlength{\tabcolsep}{5pt}
\begin{tabularx}{\textwidth}{l *{6}{>{\centering\arraybackslash}X}}
\toprule
\textbf{Method} & \textbf{Top-1 (\%)} & \textbf{Top-3 (\%)} & \textbf{Gain} & \textbf{rGain@3} & \textbf{rGain@5} & \textbf{Cov@90 (\%)} \\
\midrule
\textbf{BA-Future-JEPA (ours)} & \textbf{22.9$\pm$1.2} & \textbf{47.3$\pm$1.3} & \textbf{0.327$\pm$0.011} & \textbf{0.594$\pm$0.012} & \textbf{0.736$\pm$0.018} & \textbf{48.8$\pm$1.4} \\
Future-JEPA (generic)        & 13.0$\pm$2.6 & 35.5$\pm$3.6 & 0.213$\pm$0.030 & 0.470$\pm$0.036 & 0.629$\pm$0.036 & 36.7$\pm$3.5 \\
MAE                          & 13.5$\pm$2.1 & 31.9$\pm$1.7 & 0.201$\pm$0.026 & 0.419$\pm$0.018 & 0.566$\pm$0.019 & 32.9$\pm$1.9 \\
SimCLR                       & 11.0$\pm$1.3 & 29.3$\pm$1.4 & 0.173$\pm$0.014 & 0.400$\pm$0.017 & 0.540$\pm$0.013 & 30.4$\pm$1.3 \\
Supervised scratch           &  7.9$\pm$1.6 & 17.8$\pm$1.9 & 0.125$\pm$0.017 & 0.266$\pm$0.020 & 0.363$\pm$0.023 & 18.6$\pm$2.0 \\
\midrule
\multicolumn{7}{l}{\textit{Non-learned operational baselines}}\\
Last-best beam               & 14.7$\pm$0.5 & 24.3$\pm$0.8 & 0.226$\pm$0.007 & 0.349$\pm$0.008 & 0.403$\pm$0.009 & 25.1$\pm$0.8 \\
History-mean beam            & 11.4$\pm$0.5 & 27.3$\pm$0.7 & 0.181$\pm$0.005 & 0.376$\pm$0.006 & 0.504$\pm$0.006 & 28.3$\pm$0.8 \\
Random beam                  &  3.0$\pm$0.4 &  8.9$\pm$0.7 & 0.056$\pm$0.005 & 0.155$\pm$0.008 & 0.245$\pm$0.007 &  9.3$\pm$0.8 \\
\bottomrule
\end{tabularx}
\end{table*}

To complement the architectural discussion, we present an illustrative proof of concept for a representative 6G beam-management task, addressing a central design question: how should the JEPA target be chosen so that the learned latent representation is useful for wireless control? We consider a mmWave downlink with a $32$-antenna uniform linear array (ULA) and a $32$-beam codebook, driven by a clustered geometric channel ($3$--$6$ paths) with angular drift and random dominant-path blockage; observations are noisy log beam-power scans with random beam, time, and sector drops, while downstream labels and gain use the clean future beam-power profile. Each sample observes $8$ noisy scans; during SSL pretraining, the target branch receives the future $3$ normalized scans (steps $+1$ to $+3$), whereas the downstream probe label is the single oracle best beam at step $+3$. Table~\ref{tab:case_config} summarizes the full setup, including the source/OOD tuples and the $100/500/1{,}000$ labeled trajectories used at $1/5/10\%$ labels.

We compare five learned methods and three non-learned operational baselines. The learned methods are: (i) \emph{supervised scratch}, a Transformer encoder with a beam-classification head trained from random initialization using only the available labeled subset, i.e., without any self-supervised pretraining; (ii) \emph{SimCLR} \cite{chen2020simple}, a contrastive SSL method that pulls together embeddings of two augmented views of the same beam sequence while pushing apart embeddings of different sequences; (iii) \emph{MAE} \cite{he2022masked}, a masked-reconstruction SSL method that hides part of the input sequence and trains a decoder to reconstruct the missing beam-power values; (iv) a \emph{generic Future-JEPA} \cite{Assran_2023_CVPR}, which predicts the future latent representation from the observed context; and (v) the \emph{beam-aware Future-JEPA (BA-Future-JEPA)}, our wireless-aware instantiation. As simple non-learned references, we also include three operational baselines: \emph{last-best} (reuses the most recent best beam), \emph{history-mean} (selects the beam with the highest average power over the observed history), and \emph{random} beam selection.

All SSL methods share the same encoder architecture and the same unlabeled pretraining data. BA-Future-JEPA differs from generic Future-JEPA only by an auxiliary loss: the future standardized noisy log beam-power vector is converted into a soft beam-energy distribution via a temperature-based softmax and predicted from the context embedding. This auxiliary head is used only during pretraining and then discarded. Since the two methods are otherwise identical but share the same encoder, EMA target, latent loss, masking, and regularization- their difference isolates the effect of the wireless-aware target rather than added model capacity or a different pretext task. For downstream, all SSL methods use only the frozen encoder with a lightweight beam-ranking probe. Beyond exact prediction (Top-1/Top-3 accuracy), we report the normalized beamforming gain as an operational KPI.

Beyond exact prediction, we report operational metrics, since beam management typically shortlists a few beams for a brief confirmatory sweep. The observed gains are averaged \emph{linearly}, such as rGain@$K$ is the normalized gain when the true best of the predicted Top-$K$ beams is used, and Cov@90 is the fraction of cases where sweeping the Top-3 candidates recovers $\ge 90\%$ of oracle power (sweeping $3$/$5$ of $32$ beams costs $\approx 9.4\%$/$15.6\%$ of an exhaustive sweep). For validity, the SSL corpus, labeled pool, and every test domain are independent trajectories with disjoint random streams; each sample is a complete trajectory (no sliding-window overlap across splits); normalization uses training data only; and all hyperparameters are fixed a priori, so the five-seed results involve no test-set selection. The combined OOD tuple is a compound worst-case: cell-edge SNR, strong blockage, high mobility, and sparse feedback, rather than a single impairment.

Fig.~\ref{fig:jepa_case_study} summarizes the label-efficiency results: BA-Future-JEPA consistently improves OOD Top-3 accuracy and normalized gain, especially in the low-label case, and remains robust across the isolated stress tests. Table~\ref{tab:case_results} and Fig.~\ref{fig:jepa_refined} give the operational view on combined OOD at $5\%$ labels. Although the Top-1 gain is modest ($0.327$), the candidate-list gain is much higher: sweeping the predicted Top-3 (Top-5) beams recovers $0.594$ ($0.736$) of oracle power on average and $\ge 90\%$ in $48.8\%$ of cases, at only $9.4\%$ ($15.6\%$) sweep overhead. Thus, BA-Future-JEPA is useful as a \emph{first-stage candidate generator} feeding a short confirmatory sweep, not a single-shot selector. Its gap to generic Future-JEPA (rGain@3 $0.470$) and to MAE ($0.419$) confirms that the auxiliary future beam-energy target drives the gain.

Moreover, against non-learning-based heuristics, last-best is a strong \emph{single} guess, which combined-OOD Top-1 gain ($0.226$) exceeds the generic SSL baselines, but yields a poor \emph{candidate list} (rGain@3 $0.349$, Cov@90 $25.1\%$), precisely where BA-Future-JEPA excels ($0.594$, $48.8\%$). Additionally, persistence is competitive only in near-static regimes such as ID and isolated low-SNR, where the best beam barely changes, whereas under mobility, blockage, and missing observations the wireless-aware target retains a clear lead (mobility rGain@3 $0.83$ vs.\ $0.53$); thus target-aware latent prediction helps precisely where naive persistence fails.

We define \emph{OOD} relative to the narrow labeled source domain; since the SSL corpus spans a broad randomized range, the observed robustness primarily reflects label-efficient transfer to a narrow labeled domain rather than generalization to unseen physics (i.e., the mobility/blockage stress domains mildly exceed the SSL ranges, and the combined domain applies harsher missingness). Consistently, persistence remains competitive in near-static regimes, so the benefit should be read as robustness to non-stationarity rather than a universal improvement over classical beam tracking. Accordingly, we read the study as evidence that \emph{wireless-aware target design matters for a beam-prediction JEPA}, not as validation of JEPA-enabled 6G intelligence in general; evaluation on ray-traced or measured datasets remains future work. Finally, the encoder is intentionally small ($d{=}96$, $4$ heads, $3$ layers) so a distilled variant suits gNB/DU-local or near-RT RIC inference; detailed latency/memory profiling is left to future work.

\begin{figure}[t]
\centering
\resizebox{0.85\columnwidth}{!}{%
\begin{tikzpicture}[font=\rmfamily\small, >=Latex]

\definecolor{coreblue}{RGB}{42,96,153}
\definecolor{ringblue}{RGB}{214,229,245}
\definecolor{lightgray}{RGB}{242,242,242}
\definecolor{midgray}{RGB}{120,120,120}

\def\rin{2.00}
\def\rout{4.30}
\def\rtext{3.15}

\fill[coreblue!12] (0,0) circle (1.85);
\draw[coreblue, line width=1.0pt] (0,0) circle (1.85);

\node[align=center, font=\rmfamily\bfseries\small, text=coreblue] at (0,0.30)
{JEPA-Enabled\\6G Intelligence};
\node[align=center, font=\rmfamily\fontsize{8}{9}\selectfont,
      text width=2.8cm, text=midgray] at (0,-0.55)
{Predictive SSL for sensing, control, and automation};

\foreach \start/\col in {90/ringblue, 150/lightgray, 210/ringblue,
                          270/lightgray, 330/ringblue, 390/lightgray}{
  \path[fill=\col]
    (\start:\rin) arc[start angle=\start, end angle=\start+60, radius=\rin] --
    (\start+60:\rout) arc[start angle=\start+60, end angle=\start, radius=\rout] -- cycle;
  \draw[white, line width=1.0pt]
    (\start:\rin) arc[start angle=\start, end angle=\start+60, radius=\rin] --
    (\start+60:\rout) arc[start angle=\start+60, end angle=\start, radius=\rout] -- cycle;
}
\draw[midgray, line width=0.7pt] (0,0) circle (\rin);
\draw[midgray, line width=0.7pt] (0,0) circle (\rout);

\node[align=center, text width=2.3cm,
      font=\rmfamily\bfseries\fontsize{9}{10}\selectfont] at (120:\rtext)
{Multi-Timescale \&\\Long-Horizon\\Prediction};
\node[align=center, text width=2.3cm,
      font=\rmfamily\bfseries\fontsize{9}{10}\selectfont] at (180:\rtext)
{Action-Conditioned\\World-Model\\Control};
\node[align=center, text width=2.3cm,
      font=\rmfamily\bfseries\fontsize{9}{10}\selectfont] at (240:\rtext)
{Continual,\\Federated \&\\Distributed\\Training};
\node[align=center, text width=2.3cm,
      font=\rmfamily\bfseries\fontsize{9}{10}\selectfont] at (300:\rtext)
{Uncertainty,\\Trustworthiness \&\\Safe Control};
\node[align=center, text width=2.3cm,
      font=\rmfamily\bfseries\fontsize{9}{10}\selectfont] at (0:\rtext)
{Efficiency \&\\Hardware-Aware\\Design};
\node[align=center, text width=2.3cm,
      font=\rmfamily\bfseries\fontsize{9}{10}\selectfont] at (60:\rtext)
{Datasets,\\Benchmarks \&\\Standardization};

\end{tikzpicture}%
}
\caption{Six open research directions for JEPA-enabled 6G systems, detailed in Sec.~\ref{sec:challenges}.}
\label{fig:jepa_roadmap}
\end{figure}
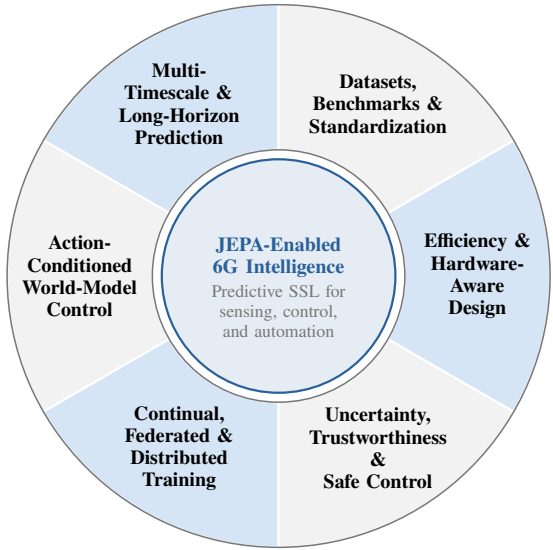

\section{Open Research Challenges and Roadmap}
\label{sec:challenges}

JEPA is a promising direction for predictive SSL in AI-native 6G networks, but its integration into operational wireless systems is still at an early stage. Moving from representation-learning prototypes to deployable 6G intelligence raises challenges in prediction horizons, action-aware modeling, distributed training, trustworthiness, efficiency, benchmarking, and standardization, as summarized in Fig.~\ref{fig:jepa_roadmap}.

\subsection{Multi-Timescale and Long-Horizon Prediction}

Wireless systems evolve across multiple time scales. Fast dynamics arise from fading, interference, bursty traffic, beam instability, and scheduling decisions, while slower variations are caused by user mobility, topology changes, service demand, and network reconfiguration. A JEPA model trained only for short-horizon prediction may capture local continuity but miss long-term dependencies needed for proactive control. Conversely, a model focused only on long-term trends may ignore short-term variations required for real-time decisions. Future JEPA designs should support hierarchical context and target windows, where short windows capture immediate CSI, beam, and KPI variations, while longer windows represent mobility, topology, and service-level evolution. Since wireless systems are uncertain and partially observable, long-horizon JEPA should also move toward uncertainty-aware and multi-hypothesis latent prediction.

\subsection{Action-Conditioned JEPA and World-Model Control}

Most JEPA formulations are observational: they predict latent targets from context but do not explicitly model how network actions affect future states \cite{lecun2022path}. However, beam switching, scheduling, handover preparation, power control, routing, and slice reconfiguration directly shape future wireless and network conditions. Without action conditioning, a JEPA model may learn correlations from traces generated by a previous control policy and may generalize poorly when the deployed policy changes. This motivates action-conditioned JEPA, where the predictor receives both the context embedding and an encoding of the action or policy applied between the context and target windows. Such a model could serve as a latent world model, allowing the network to reason about how candidate decisions affect future link quality, congestion, reliability, or mobility states. A key challenge is defining action abstractions across layers and time scales, from fast beam or MCS selection to slower cell activation, traffic steering, and slice orchestration.

\subsection{Continual, Federated, and Distributed JEPA Training}

Operational 6G networks will generate data continuously and experience non-stationary propagation, traffic, mobility, interference, and policy conditions. A JEPA model trained once on static traces is unlikely to remain reliable over time. Continual learning is therefore necessary, but online updates can introduce representation drift or catastrophic forgetting. Federated and distributed JEPA training can address privacy and scalability constraints by allowing UEs, RAN nodes, edge servers, clouds, or operators to update shared encoders without centralizing raw data. However, this introduces challenges in synchronization, communication overhead, model heterogeneity, and latent-space alignment. A specific challenge is maintaining consistent target-encoder representations across clients; if local EMA target encoders drift under different channel, traffic, or modality conditions, global aggregation may degrade. Future work should investigate representation regularization, hierarchical aggregation, domain personalization, and latent-space alignment for distributed JEPA models.

\subsection{Uncertainty, Trustworthiness, and Safe Control}

JEPA-based 6G automation must be reliable, especially for mobility management, O-RAN closed-loop control, ISAC-assisted operation, and resilient connectivity. A compact embedding is insufficient if the downstream controller cannot determine whether the prediction is reliable, ambiguous, or OOD. Therefore, JEPA models should incorporate uncertainty estimation and OOD awareness. Promising directions include probabilistic predictors, ensemble consistency, conformal calibration, latent density modeling, and prediction-residual monitoring. Trustworthy deployment also requires interpretable interfaces that allow xApps, rApps, and policy engines to reason about confidence, anomaly risk, and fallback conditions. This is essential for standards-aligned and safety-aware 6G control loops.

\subsection{Efficiency, Hardware Design, and Model Compression}

JEPA deployment will be constrained by latency, energy, memory, and communication overhead, especially in near-real-time O-RAN and edge-intelligence settings. Large multimodal encoders may provide strong representations, but cannot always satisfy RAN control-loop budgets. Efficiency must therefore be addressed at multiple levels. Tokenization choices determine sequence length and attention complexity, while encoder depth, predictor size, and latent dimensionality determine inference cost. Deployment may also require partitioning models across UEs, gNBs, edge servers, and cloud resources. Future research should explore pruning, quantization, token sparsification, early-exit inference, teacher-student distillation, and hardware-aware model design. The key challenge is to reduce complexity without destroying the predictive semantics learned during JEPA pretraining.

\subsection{Datasets, Benchmarks, and Standardization}

Progress in JEPA-enabled 6G systems is limited by the lack of common datasets and benchmarks for predictive latent learning. Existing wireless datasets are often task-specific and supervised, focusing on modulation classification, channel estimation, or beam selection. They rarely support evaluation of multimodal transfer, masking robustness, long-horizon prediction, cross-domain generalization, or operational control gains. Future benchmarks should include diverse modalities such as I/Q samples, CSI, beam measurements, KPI telemetry, topology graphs, mobility traces, and sensing observations. They should also evaluate transferability across sites, frequency bands, hardware platforms, operators, and deployment conditions.

Benchmarking should go beyond representation loss and include network-level outcomes such as beamforming gain, outage probability, BLER, latency, energy efficiency, handover failure rate, SLA violation probability, anomaly detection delay, and control-loop stability. Realistic testbeds, digital twins, and hardware-in-the-loop platforms are needed to connect representation quality with operational performance. Standardization is also important if JEPA embeddings are to be exchanged across multi-vendor 6G systems. The community should define token semantics, embedding interfaces, model lifecycle procedures, privacy requirements, drift monitoring, and interoperability mechanisms across 3GPP, O-RAN, and related 6G initiatives.

\section{Conclusion}
This article presented JEPA as a predictive SSL paradigm for AI-native 6G networks. We showed how heterogeneous wireless and network observations, including CSI, beam measurements, KPIs, topology, and ISAC data, can be tokenized and masked to train a predictive representation encoder that is reused across RAN, edge, core, and O-RAN functions through task-specific heads, rather than replacing conventional algorithms. We illustrated this reuse across beam management, link adaptation, ISAC, traffic and O-RAN automation, and security. To make the central principle concrete, an illustrative beam-management case study showed that JEPA gains depend strongly on wireless-aware target design: an auxiliary future beam-energy target during self-supervised pretraining improved label efficiency and robustness under distribution shift, relative to generic latent prediction, MAE, SimCLR, and supervised scratch baselines. Since BA-Future-JEPA and generic Future-JEPA are otherwise identical, this isolates the effect of target design rather than added model capacity. Realizing JEPA's strength in operational networks requires designing context views, target embeddings, prediction horizons, and deployment interfaces aligned with real control objectives, alongside progress in long-horizon prediction, action-conditioned modeling, distributed training, uncertainty awareness, efficient implementation, benchmarking, and standardization.



\bibliographystyle{IEEEtran}
\bibliography{ref}

\section*{Biographies}
\footnotesize

\noindent\textbf{Sheikh Salman Hassan} is a Research Associate with IDCoM, University of Edinburgh, UK. His research interests are 6G, NTN, semantic communication, and AI-native systems.

\vspace{0.1cm}
\noindent\textbf{Irshad A. Meer} is a Postdoctoral Researcher at KTH Royal Institute of Technology, Sweden. His research interests are machine learning for wireless communications.

\vspace{0.1cm}
\noindent\textbf{Almoatssimbillah Saifaldawla} is a Postdoctoral Researcher with the SIGCOM Group, SnT, University of Luxembourg. His research interests are wireless communications, resource management, and machine learning for satellite systems.

\vspace{0.1cm}
\noindent\textbf{Yan Kyaw Tun} is an Assistant Professor at Aalborg University, Denmark. His research interests are 6G networks, AI-native wireless systems, and NTN.

\vspace{0.1cm}
\noindent\textbf{Mustafa Ozger} is an Assistant Professor at Aalborg University, Denmark. His research interests are aerial and vehicular networks, IoT, and AI-assisted wireless communications.

\vspace{0.1cm}
\noindent\textbf{Madyan Alsenwi} is a Research Scientist at SnT, University of Luxembourg. His research interests are 6G networks, non-terrestrial networks, and machine learning.

\vspace{0.1cm}
\noindent\textbf{Nguyen Van Huynh} is a Lecturer at the University of Liverpool, UK. His research interests are cybersecurity, mobile computing, 5G/6G, IoT, and machine learning.

\vspace{0.1cm}
\noindent\textbf{Woong-Hee Lee} is an Assistant Professor at Dongguk University, Seoul. His research interests are signal processing, machine learning, and game theory for communications.

\vspace{0.1cm}
\noindent\textbf{Cedomir Stefanovic} is a full Professor at Aalborg University, Denmark. His research interests are non-terrestrial networks and AI-assisted wireless communications.

\vspace{0.1cm}
\noindent\textbf{Mathini Sellathurai} is a Professor at Heriot-Watt University, Edinburgh, UK. Her research interests are AI, signal processing, communications, radar, and robotics.

\vspace{0.1cm}
\noindent\textbf{Henk Wymeersch} is a Professor in Communication Systems with the Department of Electrical Engineering, Chalmers University of Technology, Sweden. His research interests are communication systems, localization, and integrated sensing and communications.

\vspace{0.1cm}
\noindent\textbf{Tharmalingam Ratnarajah} is the Fred Harris Endowed Chair in Digital Signal Processing at San Diego State University, USA. His research interests are signal processing, information theory, and 6G wireless systems.

\end{document}